\pdfoutput=1
\documentclass[11pt]{article}

\usepackage[final]{acl}

\usepackage{times}
\usepackage{latexsym}
\usepackage{amsmath}
\usepackage{esvect}

\usepackage{graphicx}

\usepackage[T1]{fontenc}
\usepackage[utf8]{inputenc}

\usepackage{microtype}

\title{Artificial Intuition: Efficient Classification of Scientific Abstracts}

\author{
         Harsh Sakhrani, Naseela Pervez, 
         Anirudh Ravi Kumar, Fred Morstatter\\
         Information Sciences Institute, Viterbi School of Engineering, University of Southern California \\ 
         {\bf Alexandra Graddy Reed} \\ 
         Sol Price School of Public Policy, University of Southern California \\
         {\bf Andrea Belz} \\ 
         Information Sciences Institute, Viterbi School of Engineering, University of Southern California
         }
\begin{document}
%\maketitle
{\makeatletter\acl@finalcopytrue
  \maketitle
}

\begin{abstract}
It is desirable to coarsely classify short scientific texts, such as grant or publication abstracts, for strategic insight or research portfolio management. These texts efficiently transmit dense information to experts possessing a rich body of knowledge to aid interpretation.  Yet this task is remarkably difficult to automate because of brevity and the absence of context. To address this gap, we have developed a novel approach to generate and appropriately assign coarse domain-specific labels. We show that a Large Language Model (LLM) can provide metadata essential to the task, in a process akin to the augmentation of supplemental knowledge representing human intuition, and propose a workflow. As a pilot study, we use a corpus of award abstracts from the National Aeronautics and Space Administration (NASA).  We develop new assessment tools in concert with established performance metrics. 
\end{abstract}

\section{Introduction}
Analyzing technical documents is a crucial strategic task, enabling the management of research portfolios, tracking investment trends, and exploring scientific advancement. On a more tactical level, it can aid the preliminary screening of scientific abstracts in systematic reviews \citep{Buchlak:2019, Rios:2015, amba:2020}. 

Several approaches are possible.  First, authors can label their own work, but this presents several challenges:  (1) authors that self-label their own texts may make idiosyncratic decisions, (2) authors in close disciplines may use different terms for related concepts, such as ``robotics'' and ``autonomy'', and (3) multidisciplinary projects may require novel or multiple labels. 

A second method is to impose an external ontology. However, these schemes often have both fine- and coarse-granularities (e.g., ``networks'' versus ``ad-hoc networks''). Another concern is that the scheme simply lacks appropriate labels, especially for emerging fields. 
% Likewise, an expert can label the documents manually, but an annotation process lacking structure and guidance can introduce noise and inaccuracies, further exacerbating the problem.  

Automated processes do exist. Those with a large number of parameters are now customizable at lower computational cost \citep{lora-2021, ben-zaken-etal-2022-bitfit}. Although dedicated pre-trained models can yield robust results, they incur significant expenses in manual annotation due to reliance on supervised learning \citep{beltagy2019scibert, chang2008importance, cohan2020specter}. 

In summary, we face two distinct needs in the analysis of scientific documents: (1) a unified, coarse-grained, non-overlapping taxonomy, tailored to uniquely classify a set of documents; and (2) an unsupervised methodology that circumvents the reliance on manual annotation while effectively managing the peculiarities of scientific text.  These challenges are exacerbated for abstracts.

In manual labeling, an expert's rapid progress often hinges on integrating prior knowledge, crucial for effective comprehension \cite{smith2021}. In so doing, the expert rapidly identifies the phrases conveying the most information and uses those for classification.  Importantly, this process is not a simple frequency or statistical analysis; indeed, the most important phrase may appear only once.  Moreover, multigrams carrying high semantic value may not appear systematically in the same place in a sentence or paragraph.  

Here we describe ``artificial intuition,'' a method mimicking the expert's process to execute two objectives: generating an optimal label space and producing accurate predictions within this new space. We integrate tools into a novel workflow to identify important terms, augment them with relevant background information, then aggregate these enhanced documents into clusters for classification purposes.

As a pilot case to evaluate our methodology, we analyze award abstracts of federally funded projects from the National Aeronautics and Space Administration (NASA) Small Business Innovation Research (SBIR) Program.  We obtain domain knowledge by extracting and ranking the abstract's keywords / keyphrases (which we will collectively refer to as ``keywords''). We generate metadata for these keywords in a zero-shot setting and derive embeddings for the keyword-metadata concatenations using a pre-trained Sentence Transformer.

For label space generation, we implement a clustering process that represents the task of organizing awards into funding themes. This method not only clarifies the thematic organization of the documents but also reveals the hierarchical relationships between different topics. We introduce a novel evaluation scheme to assess whether the label set comprehensively spans the document space and can serve as a set of basis vectors.

To predict labels, we reinterpret the multilabel-classification problem as a semantic matching challenge wherein the document space is characterized by the keyword-metadata concatenation and the label space is described by the element closest to the centroid for each cluster. This retrieval-based perspective allows for flexibility in adapting to new label spaces without the need for retraining. 

This framework accommodates various levels of parsimony, which we explore extensively in our experiments. Finally, using our test sample, we demonstrate the efficacy of our prediction methodology and quantify the performance.

\section{Related Work}
% \textbf{Extreme Multilabel Classification}.
% Multilabel classification aims to assign multiple labels to a text document. 
Various methodologies have been proposed for text classification.   Bayesian approaches \citep{tang:2016} classify the text by extracting features. One method is to first select document features with discriminative power, then compute the semantic similarity between features and documents \citep{zong:2015}, but this becomes more difficult as the number of features grows. Support Vector Machines (SVMs) can be used for document classification  \citep{cai-2004}. However, these approaches are constrained by the requirement for manual feature engineering, limiting their ability to capture the complexity of natural language.

New deep learning techniques have  advanced scientific document classification. Neural network-based architectures \citep{dernoncourt2017sequential}, particularly Convolutional Neural Networks (CNNs) \citep{cnn-2019} and Recurrent Neural Networks (RNNs) \citep{meshnet-2019, liu2016recurrent}, outperform some traditional machine learning methods. These models automatically learn feature representations from data, capturing both the semantic and syntactic nuances of text.  

These methods presume that documents are related to only one label. Newer approaches (e.g., \cite{xmtc-2017, legal-multilabel-2022, xiao-etal-2019-label,clinical-2019, chang2020taming}) classify documents with multiple labels, and one alternative attempts to map 10,000 fine-grained labels for scientific documents \citep{micol-2022} although most methods consider 10-50 coarse labels. These models are incompletely validated because many real-world datasets will have limited or poorly labeled data.

% Weakly supervised text classification methods aim to classify documents without the use of human annotated training data [CITE - Weakly supervised paper]. 
% ConWea [29] uses BERT to disambiguate the provided
% keywords and retrieve more category-indicative words for pseudo training data collection; LOTClass [33] leverages one BERT encoder to perform masked language modeling for finding more indicative words and another BERT to perform classification.

Weakly supervised learning and zero-shot learning (ZSL) models do not use annotated data. Some pre-trained language models demonstrate impressive performance in zero-shot document classification \citep{bert-2019, beltagy2019scibert, roberta-2019} and can be used to assign multiple labels to a given document \citep{zero-shot-2019}. On the other hand, hierarchical multi-class methods can use just class names - without training examples - as supervision \citep{taxoclass-2021, metadatainduced-2022}.  Large language models trained on scientific data, such as Galactica \citep{galactica-2022} and SciNCL \citep{scincl-2022}, can be used to assign labels to a scientific document.  Many approaches use metadata, such as generic descriptions, as supervision for further classification  \citep{zhang-futex-2023}.  However, these methods are still potentially subject to noise.  Here we describe a method to identify keywords and derive context-specific metadata to improve classification accuracy, particularly for short abstracts.  

%However, these approaches do not address the critical information embedded within extraneous content nor do they take into account the context-specific metadata. 

\section{Approach}
\subsection{Problem Formulation}
The scientific literature tagging task can be conceptualized as a multi-label classification (each paper can be relevant to more than one label) problem, where all candidate tags (e.g., ``Aerodynamics,'' ``Superconductance/Magnetics'') constitute a label space $Y$ of arbitrary size.  We seek to: 

\begin{itemize}
\item Construct a new label space $Y$ comprising coarse-grained labels and aggregating correlated labels  (e.g., merging ``Optics'' and ``Photonics'' into ``Optical technologies'').
\item Develop an unsupervised multi-label classifier that can effectively map an abstract to the new label space $Y$.
\end{itemize}

A simplistic approach would utilize a pre-trained language model to encode each document and label, generate their embeddings, and then conduct a nearest neighbor search in the embedding space. However, this method encounters two primary challenges: (1) the existing language models are largely trained on general English text that does not discern technical terms, and (2) analogous labels (e.g., ``networking'' and ``ad-hoc networks'') confound the results. One might augment the label embedding process with generic metadata, such as a brief description from Wikipedia or using solutions like Positive Instance Feature Aggregation (PIFA) \cite{yu2022pecos}.

Instead, we seek to generate a context-specific glossary.  This has the added advantage that the labels can be fine-tuned, converting a multi-label problem into a simpler system.  For instance, a thermal protection system (TPS) consists of materials suited to handle extremely high temperatures.  In a conventional classification scheme, this might require two labels, such as ``materials'' and ``temperature.''  In contrast, we create a system such that ``thermal protection system'' is itself sufficient to serve as the only label.  This is possible only with a label space customized to the knowledge domain. %In other words, the parsimony of the classification problem is a function of how accurately the constructed label space conforms to the to the peculiarities and the specifics of the knowledge domain.

\subsection{Implementation Components}
\begin{itemize}
    \item Yet Another Keyword Extractor (YAKE) \citep{CAMPOS2020257} is a lightweight, unsupervised keyword extraction algorithm that uses statistical properties and contextual information.
    
    \item Mistral 7B is a Large Language Model (LLM) with strong performance of Llama-2 13B on key benchmarks \cite{jiang2023mistral}.

    \item Maximal Marginal Relevance (MMR) \citep{mmr-1998} iteratively selects candidate items that simultaneously maximize their relevance to the query and their novelty compared to previously selected items.

% Figure 1 
\begin{figure*}
    \centering
    \includegraphics[width=\textwidth]{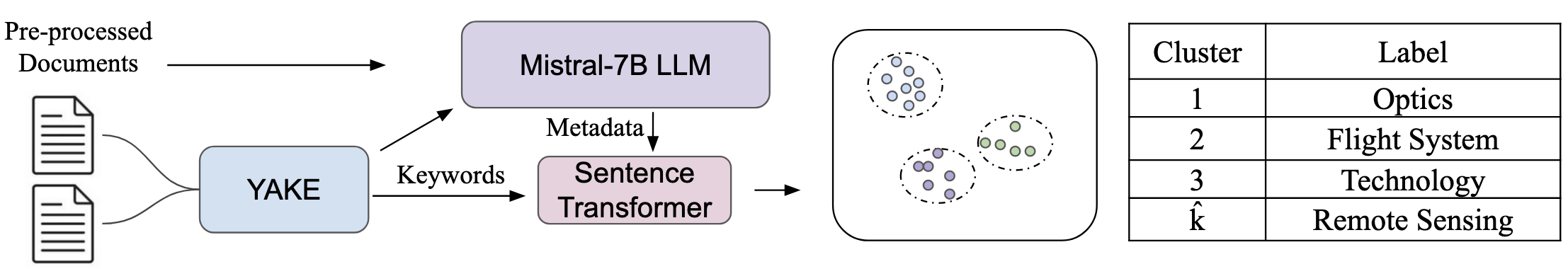}
    \caption{Label Space Generation flowchart.  The clusters are named with the keyword closest to the cluster centroid.}
    \label{fig:label-gen}
\end{figure*}

\item Sentence Transformer (S-Transformer)\footnote{https://huggingface.co/sentence-transformers/all-mpnet-base-v2} \cite{reimers2019sentencebert} constructs dense vector representations of sentences to enable efficient comparison of text semantics.

\end{itemize}

\subsection{Document Corpus}
\label{subsec:document-corpus}

The NASA SBIR program publishes abstracts of funded projects. We used 1,230 abstracts from 2010 to 2015 extracted online from the publicly available archive\footnote{sbir.gov}.  The average abstract length is about 450 words.  All abstracts were pre-processed by removing stop words, which were variations and combinations of: ``NASA'', ``space'', ``mission(s)'', ``research'', ``SBIR'', ``spacecraft'', ``future'', and ``science''.  These words and multigrams comprising these words appeared in a large number of the abstracts, and therefore they provided little information to assist in classification. We randomly drew 100 abstracts 
(roughly 10\%) for manual classification, described below.  
\subsection{Label Space Generation}
\label{sec:generation}

\begin{figure}
    \centering
    \includegraphics[width=0.5\textwidth]{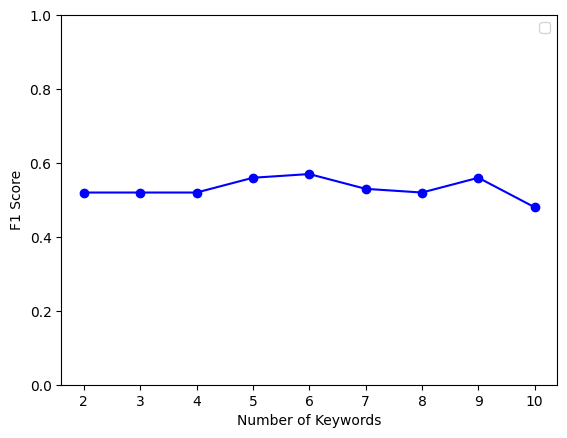}
    \caption{Variation of F1 score with the number of keywords at the threshold of top 1\%.}
    \label{fig:keywords} 
\end{figure}

We generate the label space as illustrated in Figure \ref{fig:label-gen}. Initially, pre-processed abstracts are submitted to YAKE to extract keywords. One hyperparameter of our workflow is the number of keywords $\hat{c}$. We estimated that $\hat{c}$ should be approximately 5 as it represents 1-2\% of the abstract length. We confirmed that the F1 results, described in more detail below, showed a general lack of sensitivity to this parameter (Figure \ref{fig:keywords}), and therefore we set $\hat{c} = 5$ in our main analyses.

% Figure 2

We sought to supplement these keywords with contextual definitions to form metadata. We used Mistral-7B Instruct v0.2 with hyperparameters set at default values and submitted the following prompt:
\begin{quote}
    \textit{Given the scientific abstract and the keywords that have been extracted for the document, provide a concise meta data/prior information for every keyword in context of the document. Incorporate any extra knowledge that can help classify the document to relevant topics.} 
\end{quote}

 This combined data-keyword concatenation is processed using the S-Transformer model to produce embeddings. A critical aspect of this process is that the metadata generated for each keyword is tailored specifically to the context of the related document, ensuring that the embeddings are context-specific rather than generic. We use k-means clustering \citep{keyword-clustering-2015} to partition these embeddings into clusters, represented by the keyword closest to their centroids and effectively summarizing each cluster's thematic focus. This method approximates the scheme by which such abstracts would be sorted in a funding portfolio.  

Unlike $\hat{c}$, the number of clusters $\hat{k}$ requires closer examination.  We seek a parsimonious model that minimizes the number of labels per document.  In practice, we seek to organize approximately 1,000 documents into approximately 10-20 classes. In addition to making this a tractable problem, it adequately represents the portfolio management process. 

\subsection{Annotation Task Design}\label{sec:annotate}
We conducted a manual annotation task to label the test set of the NASA SBIR abstracts. We presented the annotator, a NASA expert, with a scientific abstract and the generated label set. The annotator was instructed to assign a label to the scientific abstract only if one of the presented labels was appropriate, and to leave it unlabeled otherwise. The same documents were labeled for each configuration for consistency.  

\section{Results}
\subsection{Label Space Orthogonality: Redundancy}
\label{subsec:red}

Our first task is estimate the degree of overlap within the label space.  To do so, we define the redundancy, $\mathcal{R}$, as a measure of the orthogonality between labels. This figure of merit (FOM) is intrinsic to the label space and assessed independently of individual document projections. 

The labels are transformed into normalized embeddings using the S-Transformer model, resulting in a label matrix $\mathcal{L}$ of dimensions $\hat{k} \times v$ (in our case, $v = 768$).  Each element $\mathcal{L}_{ij}$ represents the $j$-th dimension of the $i$-th label embedding. 

To measure the orthogonality, we calculate the cosine similarity between each pair of distinct label embeddings. If the labels are orthogonal and distinct, the cosine similarity should approach 0; on the other hand, two labels capturing closely related ideas will give a cosine similarity that approaches 1. Formally, for normalized label vectors $\mathcal{T}_i$ and $\mathcal{T}_j$ in $\mathcal{L}$, we define redundancy $\mathcal{R}$ as the maximum cosine similarity among all pairs:

\begin{equation}
% \[
\mathcal{R} = \max_{i \neq j} (\text{cosine similarity}(\mathcal{T}_i, \mathcal{T}_j))
% \]
\end{equation}

where

\[
\text{cosine similarity}(\mathcal{T}_i, \mathcal{T}_j) = \frac{\mathcal{T}_i \cdot \mathcal{T}_j}{\|\mathcal{T}_i\| \|\mathcal{T}_j\|}
\]

A value of $\mathcal{R}$ close to 0 is desirable because orthogonal label embeddings  suggest that each label contributes unique information without redundancy. Conversely, a value of $\mathcal{R}$ approaching 1 shows that at least one pair of labels shares a high degree of overlap. Overlap implies that multiple labels may be describing similar features within the documents, thus  complicating the interpretability and utility of the label space.  Our goal is to represent each key concept with a unique label.  

\begin{figure}
    \centering
    \includegraphics[width=0.5\textwidth]{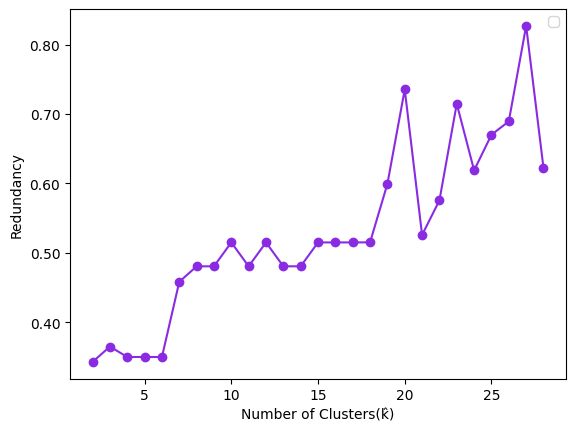}
    \caption{Variation of redundancy $\mathcal{R}$ with the number of clusters $\hat{k}$.}
    \label{fig:red}
\end{figure}

To understand the redundancy in our basis vector set, we executed the label space generation process but systematically varied $\hat{k}$.  We then evaluated $\mathcal{R}$ for each label space.  $\mathcal{R}$ increased with $\hat{k}$ (Figure \ref{fig:red}), as expected. Notably, we identified three general regimes:  At low $\hat{k}$, $\mathcal{R}$ was fairly flat and low.  The labels do not overlap.  At approximately $\hat{k} = 8$,  the redundancy increased to a new plateau.    At much higher values of $\hat{k}$ (18 and higher), this FOM entered a regime in which the value dramatically oscillated.  

We therefore conclude that at 
very low cluster numbers ($\hat{k} < 6$), the severely reduced $\mathcal{R}$ indicates that the labels are probably insufficient to describe the document set. At higher values of $\hat{k} > 18$, the risk of overlapping labels increases substantially, but the likelihood that each document is at least minimally described also increases.

% Figure 3: Redundancy plot

\subsection{Spanning the Document Space: Coverage}
\label{subsec:cov}

We defined the redundancy $\mathcal{R}$ to characterize the orthogonality of our proposed label space basis vectors.  Next, we study how comprehensively these labels describe the documents, essentially determining if our labels can span the document space. 

% Figure 4:  Analysis workflow 
\begin{figure*}
    \centering
    \includegraphics[width=\textwidth]{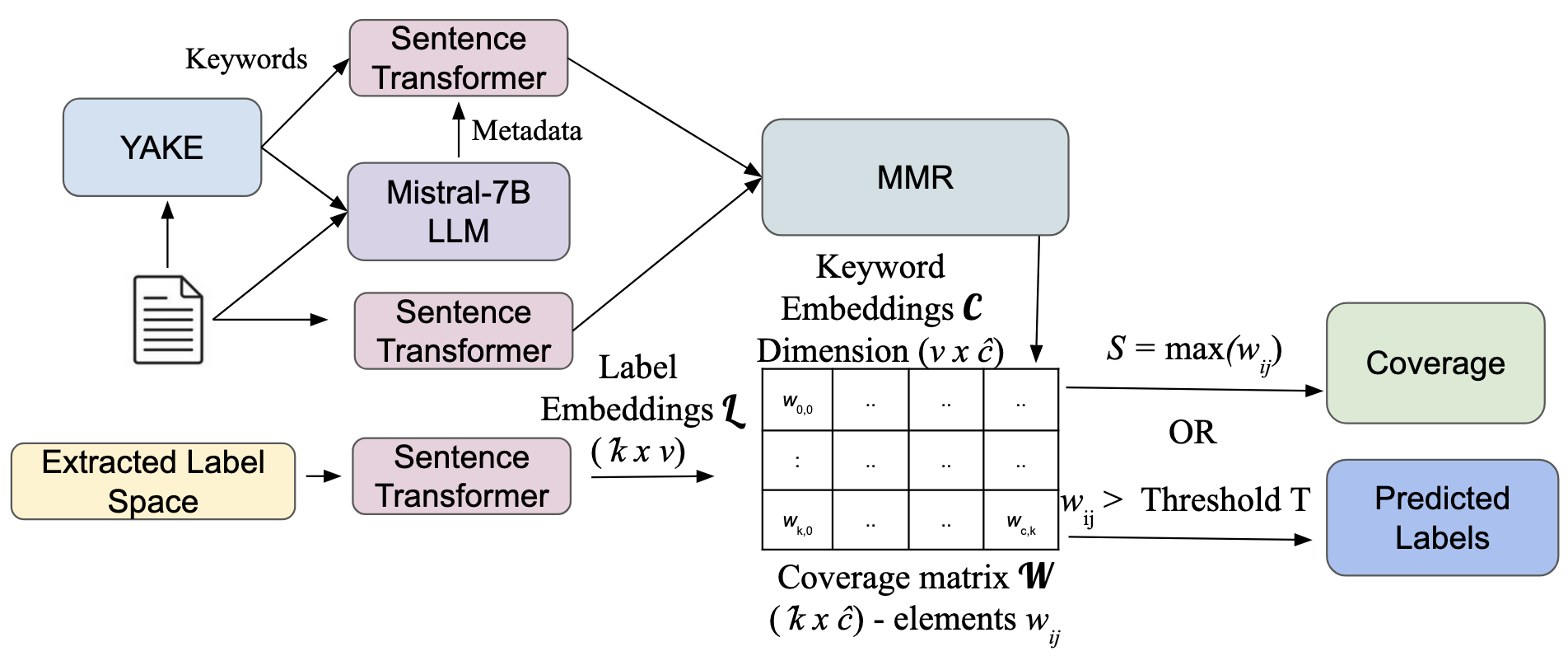}
    \caption{Analysis workflow and use of the coverage matrix $\mathcal{W}$.  In one application (final step in green), the element with the maximum value is used to generate the Coverage.  The second usage (blue final step) is to extract those values exceeding a specific threshold $T$ for the label prediction task. }
    \label{fig:label-eval}
\end{figure*}

We architected a second workflow  (Figure \ref{fig:label-eval}). Again we begin with YAKE usage for a single document. We submit these keywords to Mistral-7B for document-specific contextual definitions as supplementary metadata. Both the document itself and the keyword-metadata concatenations are subsequently processed through the S-Transformer model to generate their individual embeddings, refined using MMR.  This forms a new keyword embedding matrix $\mathcal{C}$ of dimensions $v \times \hat{c}$, where $v$ (768 in our case) represents the embedding dimension, and the extracted keywords are still parameterized by $\hat{c}$. 

Likewise, we still have the label embedding $\mathcal{L}$ of dimensions $\hat{k} \times v$.  As our goal is to understand the overlap between the labels and the corpus embeddings, we define a new matrix, termed ``coverage'', $\mathcal{W}$ with elements $w_{ij}$: 

\begin{equation}
w_{ij} = \sum_{v}{L_{iv}C_{vj}},     
\end{equation}

The coverage matrix $\mathcal{W}$ has the resulting dimension $\hat{k}$ x $\hat{c}$, where $\hat{k}$ represents the number of labels and $\hat{c}$ represents the number of keywords.  In other words, $\mathcal{W}$ is the projection of the keywords onto the label space. (Strictly speaking, the S-Transformer embeddings of length $v$ can be understood as creating a coordinate system to facilitate projections.)   Each $w_{ij}$ element ranges from -1 to 1.  

A high value of any element $w_{ij}$ indicates that a label and keyword are highly aligned. Therefore, finding the maximum value that appears in this matrix $\mathcal{W}$ will signify how well the label space describes the keywords of an individual document in the best case.  Consequently, we define the coverage $\mathcal{S}$ for a given document $d$ (where $d$ is a member of the document corpus $\mathcal{D}$):   

\begin{equation}
\mathcal{S}^d = \textrm{max}(w^d_{ij})      
\end{equation}

The coverage for the corpus $D$ is simply the average of the individual documents' coverage: 

\begin{equation}
\mathcal{S}^D = \frac{\sum_D S^d}{D}      
\end{equation}

% Figure 5: Coverage figure
\begin{figure}
    \centering
    \includegraphics[width=0.5\textwidth]{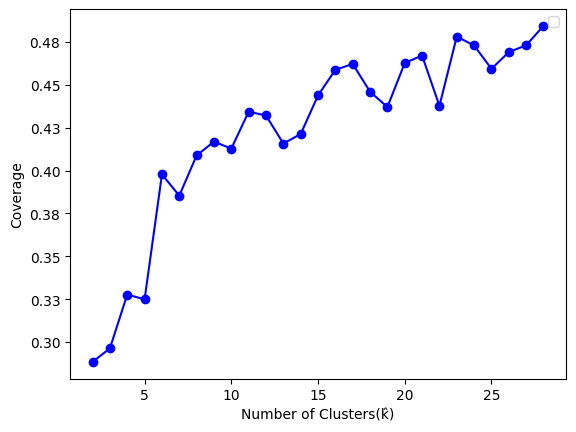}
    \caption{Variation of coverage $S$ with $\hat{k}$.}
    \label{fig:cov}
\end{figure}

This proposed figure of metric, coverage, provides critical validation that the new label space is pertinent to the knowledge domain encompassed by the documents.  One would expect for coverage to be small if the label space is not large enough - namely, for small values of $\hat{k}$.  An intermediate regime would appear if each new label adds significant new information. Eventually, a final regime would be reached wherein the new information provided by an additional label is marginal as the segmentation becomes finer, such as comparing `Chemical Propulsion Technologies' and `Electronic Propulsion Technologies'. In other words, a general analytical form for coverage should start near 0, then experience rapid growth until the space is largely covered and it tapers off.   The corpus coverage is bounded by 1 because the individual documents' coverage is given by a cosine similarity of two normalized vectors, thus limited to 1.

We tested this concept by varying the number of clusters $\hat{k}$ from 2 to 28 and evaluating coverage $\mathcal{S}$ for each newly developed label space.  As $\hat{k}$ increased, the labels did indeed relate well to the documents, as represented by keywords (Figure \ref{fig:cov}). In addition, the variation revealed a generally asymptotic form, as expected.

\begin{table}[ht]
\caption{Labels at $\hat{k}=15$}
\label{tab:k.15}
\centering
\scalebox{0.85}{
\begin{tabular}{c}
\hline
Advanced Optical Systems \\
Advanced Photovoltaic Systems \\
Aeroservoelastic Analysis and Aircraft Systems Analysis\\
Aeroservoelastic Analysis Tools \\
Electric Propulsion Systems\\
Electrolyzers \\
High Energy Density Electronics \\
LIDAR Remote Sensing \\
Multifunctional Composite Materials \\
Optical Communications Technology \\
Radiation-Hardened Electronics \\
Robotic Science Missions \\
Technologies Fault Management \\
Thermal Protection Systems \\
Unmanned Aircraft Systems \\
\hline
\end{tabular}}
\end{table}

\subsection{Label Assignment: Precision and Recall}

\begin{table}[ht]
\caption{Labels at $k=25$ }
\label{app:k=25}
\centering
\scalebox{0.75}{
\begin{tabular}{c}
\hline
Advanced Aeroservoelastic Analysis and Rotorcraft Aeromechanics \\
Advanced Composite and Ceramic Matrix Materials \\
Advanced ESR Technologies for Space Exploration \\
Advanced Energy Storage and Power Systems \\
Advanced Fluid and Thermal Management Technologies \\
Advanced Laser and Optical Communication Technologies \\
Advanced Manufacturing Technologies for Aerospace \\
Advanced Microwave and Remote Sensing Technologies \\
Advanced Optical Systems for Scientific Missions \\
Advanced Structural Sensors and NDE Technologies \\
Advanced Thermal Protection Systems \\
Airborne Measurement and Sensing Systems \\
Automation and Control in Robotic Science Missions \\
Fault Management Technologies \\
High-End Computing and Data Handling \\
Highly Capable Propulsion Systems \\
Innovative Aerospace Structural Design \\
Innovative Fiber-Optic and Navigational Technologies \\
International Space Station \\
LIDAR Remote Sensing Technologies \\
Mars Sample Return Missions \\
Radiation-Hardened Electronics and Sensors \\
Regenerative Life Support Systems \\
Solar Power Technologies for Advanced Energy Solutions \\
Unmanned Aircraft Systems Operations \\
\hline
\end{tabular}}
\end{table}

We seek to create a label space with high coverage, indicating relevance; and low redundancy or overlap.  However, these measures act in opposition as higher coverage naturally can lead to greater redundancy.  That is, these two measures form a trade space in which we strive to optimize $\hat{k}$.  

We revisited the workflow generating the coverage matrix (Figure \ref{fig:label-eval}) and developed a prediction pipeline, mirroring the initial process through the creation of the coverage matrix $\mathcal{W}$.

%The elements of this matrix, $w_{ij}$, quantify the overlap between each label and keyword as represented in a high-dimensional space. 
%A high $w_{ij}$ indicates a significant overlap between the label and the keyword. 

In the coverage study, we took the maximum $w_{ij}$ value to characterize the space.  Here, we seek to find \textit{all} relevant values of $w_{ij}$.  To operationalize this, we analyze the distribution of all $w_{ij}$ values and establish a threshold $T$, which defines the minimum percentile to be used as a filter for the $w_{ij}$ values, effectively distinguishing between significant and negligible overlaps.
For instance, setting $T = 1\%$ means we retain only the top 1\% of the $w_{ij}$ values, which is more restrictive than setting $T = 10\%$. In practical terms, for a system of 5 keywords and 15 labels, a 1\% threshold would retain just one label (top 1\% of 5x15 = 75 matrix elements results in one). On the other hand, a 10\% threshold retains seven elements that could be distributed in various ways.  For instance, all five keywords might describe label 1, with two of those keywords linked to label 2; or only one keyword could be associated with each of seven labels.  As the threshold $T$ gets larger, the variability in possible outcomes increases.

For each document, we select labels associated with the values of $w_{ij}$ that exceed the threshold $T$.  However, to accurately evaluate the classification, a set of `true' labels is required. While NASA maintains its own taxonomy of approximately 200 labels that could theoretically serve this purpose, the inconsistency in this taxonomy year-to-year and the excessive number of labels compared to our needs complicate its use. Instead, as noted in Section \ref{sec:annotate}, we manually aligned the abstracts with our new labels.

Using the three regimes of Figure \ref{fig:red} as a guide, we considered three values for $\hat{k}$ - 4, 15, and 25 - and estimated the usual classification measures of precision, recall, and F1.  Moreover, we varied the threshold $T$, hypothesizing that at low restrictive values of $T$, these measures should improve as only the most significant overlaps in the coverage matrix would be retained. 

At $\hat{k}$ = 4, the labels were: Propulsion Technologies, Remote Sensing Technologies, 
Thermal Protection Systems, and Unmanned Aircraft Systems.  However, the manual classification task failed because the labels simply did not describe the abstracts. 

At $\hat{k} = 15$, the labels consisted of words generally associated with space technologies (Table \ref{tab:k.15}). Similarly, the $\hat{k} = 25$ generated labels related to space (Table \ref{app:k=25}); however, in this case the word ``Advanced'' preceded nearly half the technical topics, suggesting that the semantic content of that word decreases in this context.    (Notably, the word ``advanced'' has been linked to other technical contexts where its semantic content is diluted \citep{Belz2023}).

\begin{figure}[ht]
    \centering
    \includegraphics[width=0.4\textwidth]{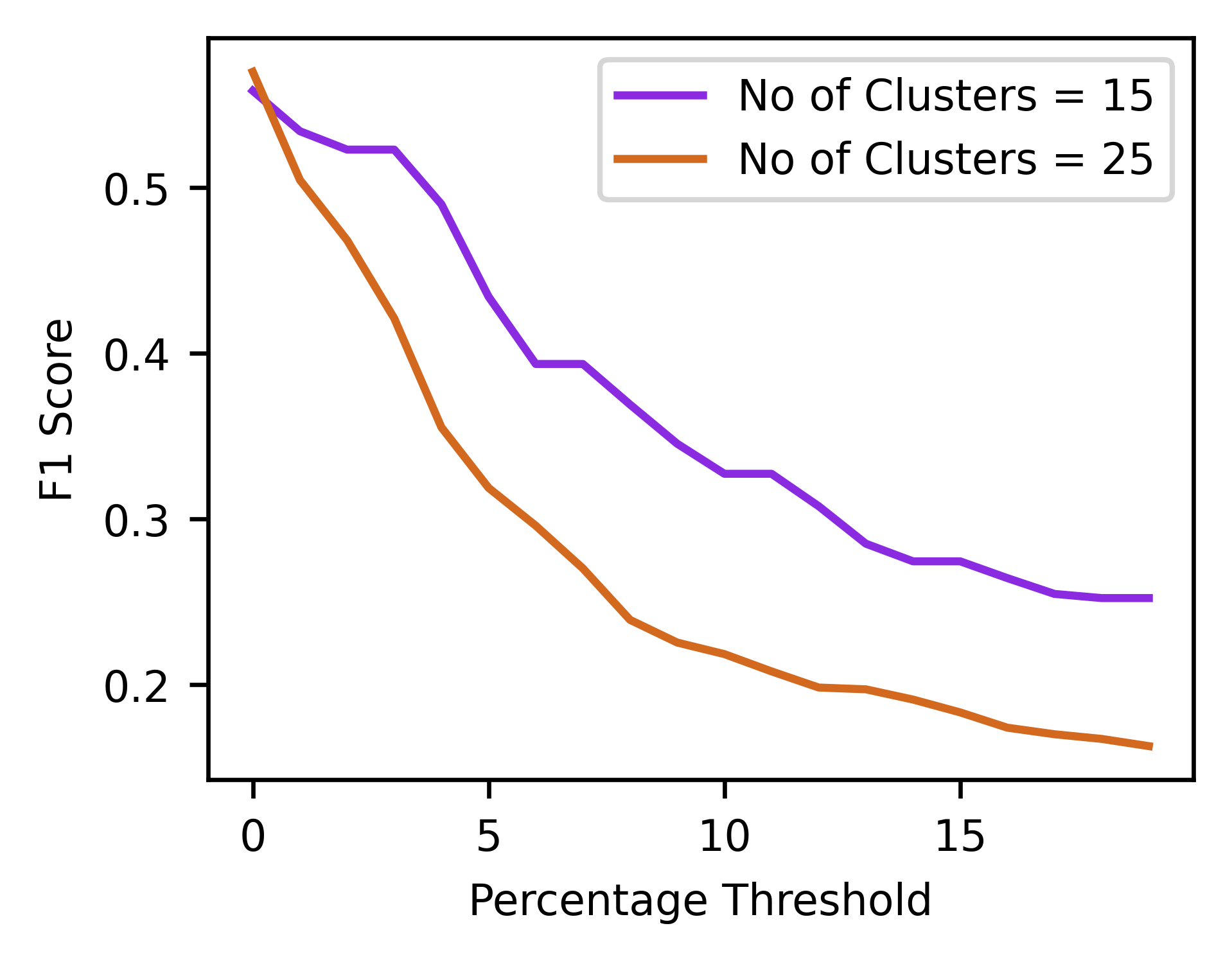}
    \caption{Variation of F1 scores for assigned labels with weights $w$ exceeding the percentile threshold $T$, as defined in the text.}
    \label{fig:k.15.vs.25}
\end{figure}

To evaluate our method's quality, we set aside $\hat{k}=4$ and considered differences between $\hat{k} = 15$ and $\hat{k} = 25$.  We focused on the F1 score and found that  $\hat{k} = 15$ consistently yielded higher scores than the overdetermined space represented by $\hat{k} = 25$ (Figure \ref{fig:k.15.vs.25}). 
As a result, we concluded that $\hat{k} = 15$ represented a better set of labels to describe this space. 

% Figure 6:  k=15 vs 25:  F1

\begin{figure}[ht]
    \centering
    \includegraphics[width=0.4\textwidth]{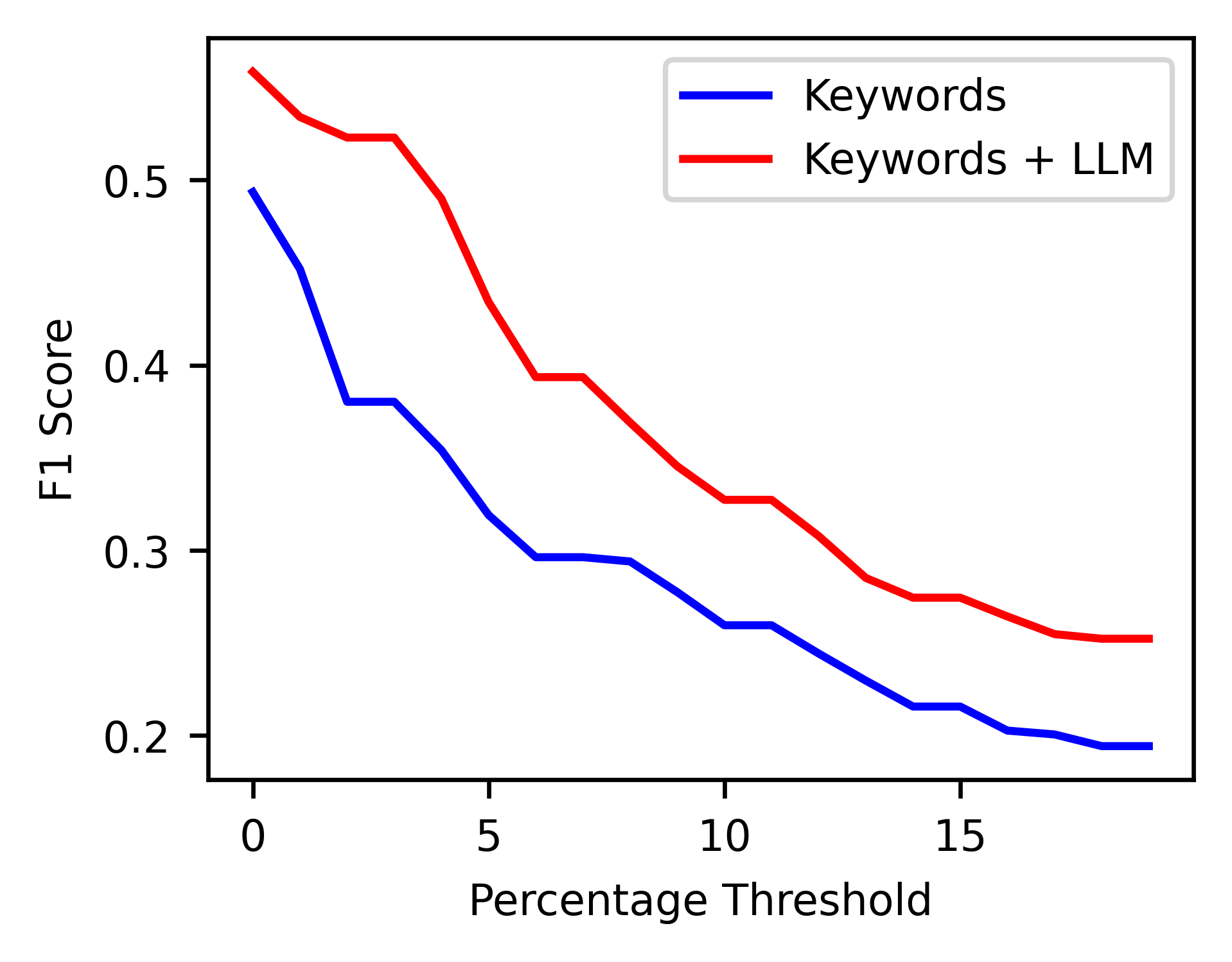}
    \caption{Variation of F1 scores for assigned labels with weights $w$ exceeding the percentile threshold $T$, as defined in the text, for $\hat{k}=15$.}
    \label{fig:LLM.impact}
\end{figure}

Our final task was to demonstrate the advantage of augmenting the abstract with the metadata extracted from the additional analysis of the keywords.  Using the $\hat{k} = 15$ label space described above, we evaluated the performance of our model with and without the metadata generated by the LLM.  We found that the LLM consistently improved the F1 score (Figure \ref{fig:LLM.impact}) for all tested values of the threshold $T$. This was due to improvement primarily in the precision (Table \ref{tab:k.15.performance}). 

% Figure 7 - with /without LLM

\begin{table}[ht]
\caption{Precision, recall and F1 scores for $k=15$ at varying thresholds ($T$).}
\label{tab:k.15.performance}
\centering
\begin{tabular}{crrr}
\hline
\textbf{$T$ (\%)} & \multicolumn{1}{c}{\textbf{Precision}}   & \multicolumn{1}{c}{\textbf{Recall}}  & \multicolumn{1}{c}{\textbf{F1}}\\
\hline
1 & 0.56 & 0.56 & 0.56 \\
5 & 0.35 & 0.79 & 0.49 \\
10 & 0.22 & 0.81 & 0.35 \\
15 & 0.16 & 0.86 & 0.27 \\
20 & 0.15 & 0.91 & 0.25 \\
\hline
\end{tabular}
\end{table}

% \newpage
\section{Discussion and Future Research}
Scientific communication is designed to efficiently carry rich information between experts.  The abstract of a grant or publication is perhaps the most striking example, wherein sophisticated concepts are conveyed in a relatively dense, short vehicle.  Years of study generate a large body of knowledge to guide the expert in a classification process.  Indeed, this additional material and the associated judgments underpin the rapid decision-making characteristic of human intuition. 

We have sought to replicate that process in an automated methodology.  Our unsupervised approach is robust and flexible, enabling its use in various domains. Its independence from specific label sets underscores its adaptability and broad applicability.  Our contributions range from applied text processing tasks to economics and public policy, with several interesting directions ahead. 

First, we have tested this approach on a relatively narrow set of abstracts by selecting a NASA corpus of documents as the first test case.  This exercise should be conducted on benchmark datasets such as Maple (Metadata-Aware Paper colLEction) \footnote{https://github.com/yuzhimanhua/MAPLE/tree/master/}. This would demonstrate the generalizability of our approach.

% narrow fields, such as abstracts of grants issued by the National Institutes of Health; or on a broader set, like those of the National Science Foundation. 
% Effectively, this would generate corpora with varying cluster distances, which could impact the efficacy of the methods proposed here. 

Second, a different validation would be to compare these results with those of longer documents.  For instance, one could analyze both publication abstracts and the full text.  It is not clear if the publications would contain more noise; or perhaps the complete text would carry the metadata such that the LLM task would be less necessary. 

In addition, here the manual classification exercise assigned only one label to each abstract as a rigorous test. We have not explored the opportunity to generate multiple labels for a single abstract. Indeed, the $\hat{k}=25$ data set points to this, as some of the labels (such as ``Advanced Thermal Protection Systems'') addressed the technology itself, while others described the intended application (e.g., ``Mars Sample Return missions'').  In the future, we can develop a new weighting scheme addressing this complex classification.

Finally, our method opens lines of inquiry in business or public policy, as we could use this labeling method to generate metadata for the abstracts themselves.  In this fashion, the labels could form a variable to be used in further assessment, such as patterns in funding, research direction, patents, or other corpora where scientific documents are condensed in short summaries.  Using this method with public company reports could create entirely new industry categories, updating existing schemes \citep{Shweta2021, Hoberg2010, Hoberg2016}.  Moreover, these data could be combined with other tags, such as principal investigator, institution, or other bibliometric characteristics to create a complex profile. Such a data set could be used to track a number of interesting trends.

\section{Conclusion}
For labeling short scientific documents, such as abstracts, pre-existing domain-specific taxonomies are ambiguous. Defining a label space spanning the set of documents is an important task that humans execute easily. In this paper, we demonstrate that the text of the documents is insufficient to either define the label space or predict the labels. We present evidence that an LLM can provide critical metadata to address this gap, forming the basis for artificial intuition. Additionally, we propose both an architecture to address this and two novel measures to evaluate the constructed label spaces. Testing our model with a corpus of NASA award abstracts, we demonstrate a workflow that integrates the LLM's supplemental data successfully.

%\section*{Acknowledgments}

%We thank the Management of INnovation, Entrepreneurial Research, and Venture Analysis (MINERVA) group and the National Academies Review of the Small Business Innovation Research and Small Business Technology Transfer Programs at the Department of Defense.

\newpage
\nocite{Fortunato:2018, Buchlak:2019, Rios:2015, amba:2020,
beltagy2019scibert, chang2008importance, 
cohan2020specter, banerjee2020segmenting, 
dernoncourt2017sequential, tang:2016, zong:2015, cai-2004, cnn-2019, weak-supervised-2023}

\bibliography{anthology,custom}
\bibliographystyle{acl_natbib}

%appendix

\end{document}